\documentclass[10pt,twocolumn,letterpaper]{article}
\usepackage{iccv}
\usepackage{times}
\usepackage{epsfig}
\usepackage{graphicx}
\usepackage{amsmath}
\usepackage{amssymb}
\usepackage[linesnumbered,algoruled,boxed,lined]{algorithm2e}
\usepackage{multirow}
\usepackage[table,xcdraw,dvipsnames]{xcolor}
\usepackage{xcolor}
\usepackage{utfsym}
\definecolor{FloralWhite}{RGB}{51,51,225}
\definecolor{Soft_Blue}{RGB}{0,153,0}
\usepackage[accsupp]{axessibility} 

\usepackage[breaklinks=true,bookmarks=false]{hyperref}

\iccvfinalcopy 

\def\iccvPaperID{6193} 

\ificcvfinal\fi

\begin{document}

\title{Boosting Few-shot Action Recognition with Graph-guided Hybrid Matching}

\author{
Jiazheng Xing$^{1}$\thanks{Equal Contribution.},
~Mengmeng Wang$^{1}$\footnotemark[1],
~Yudi Ruan$^{1}$,
~Bofan Chen$^{1}$,
~Yaowei Guo$^{1}$,\\
~Boyu Mu$^{1}$,
~Guang Dai$^{2,3}$,
~Jingdong Wang$^{4}$,
~Yong Liu$^{1}$\thanks{Corresponding author.}
\vspace{0.8em}\\
$^{1}$ Zhejiang University, $^{2}$ SGIT AI Lab,  $^{3}$ State Grid Corporation of China, $^{4}$ Baidu Inc.\\
\tt\small \{jiazhengxing, mengmengwang, yudiruan, bofanchen, guoyaowei, muboyu\}@zju.edu.cn\\
\tt\small \ yongliu@iipc.zju.edu.cn, guang.gdai@gmail.com, wangjingdong@baidu.com
}
\maketitle
\ificcvfinal\thispagestyle{empty}\fi

\begin{abstract}
   Class prototype construction and matching are core aspects of few-shot action recognition. Previous methods mainly focus on designing spatiotemporal relation modeling modules or complex temporal alignment algorithms. Despite the promising results, they ignored the value of class prototype construction and matching, leading to unsatisfactory performance in recognizing similar categories in every task. In this paper, we propose GgHM, a new framework with \textbf{G}raph-\textbf{g}uided \textbf{H}ybrid \textbf{M}atching. Concretely, we learn task-oriented features by the guidance of a graph neural network during class prototype construction, optimizing the intra- and inter-class feature correlation explicitly. Next, we design a hybrid matching strategy, combining frame-level and tuple-level matching to classify videos with multivariate styles. We additionally propose a learnable dense temporal modeling module to enhance the video feature temporal representation to build a more solid foundation for the matching process. GgHM shows consistent improvements over other challenging baselines on several few-shot datasets, demonstrating the effectiveness of our method. The code will be publicly available at \href{https://github.com/jiazheng-xing/GgHM}{https://github.com/jiazheng-xing/GgHM}.
\end{abstract}
\vspace{-5pt}
\section{Introduction}
\begin{figure} [ht]
		\centering
		\includegraphics[width=\linewidth,height=1.065\linewidth]{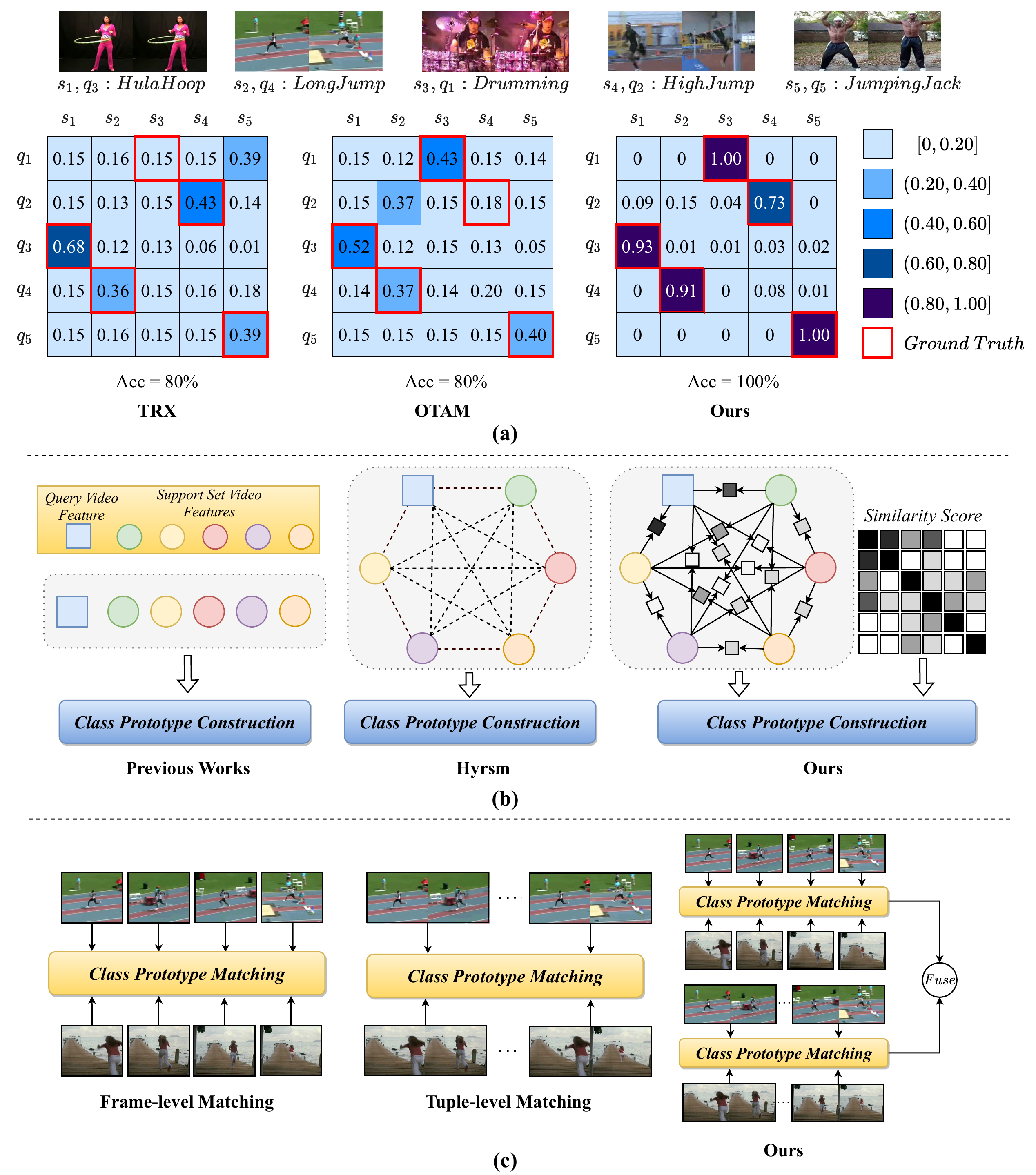}
		\caption{ (a): Similarity visualization between query and support videos with different methods on the 5-way 1-shot task of UCF101~\cite{soomro2012ucf101}. A higher score indicates a greater degree of similarity. TRX~\cite{perrett2021temporal} misclassifies the drumming as the jumping jack, and OTAM~\cite{cao2020few} misidentifies the high jump as the long jump. Our method identifies all categories of videos accurately. (b): Different types of class prototype construction. Previous works did not do any information interaction among different videos. HyRSM~\cite{wang2022hybrid} operates an inter-relation function without leveraging label-informed supervision. Our method utilizes the graph network with label-informed supervision to learn the correlation between different videos. 
  (c): Different types of class prototype matching. Frame-level matching~\cite{zhu2018compound, cao2020few, wang2022hybrid} uses single individual frames for matching, while tuple-level~\cite{thatipelli2022spatio, xing2023revisiting, perrett2021temporal} matching combines several frames into a tuple as the matching unit. Our method combines both to complement each other's shortcomings. }  
	\label{fig:fig1}
	\end{figure}
Compared with general action recognition, few-shot action recognition requires limited labeled samples to learn new categories quickly. It can avoid the massive, time-consuming, and labor-consuming data annotation commonly associated with supervised tasks, making it more adaptable for industrial applications.
According to this advantage, increasing attention has been directed toward the field of few-shot action recognition~\cite{cao2020few, perrett2021temporal, thatipelli2022spatio, wang2022hybrid, zheng2022few, huang2022compound, nguyen2022inductive, xing2023revisiting}. However, since few-shot action recognition has limited learning material, learning well-generalized models are challenging.

 Current attempts to address the above problems ~\cite{cao2020few, zhu2018compound, zhang2020few, perrett2021temporal, thatipelli2022spatio, xing2023revisiting, wang2022hybrid} mainly adopt the metric-based framework and episode training to solve the difficulty of model migration on new categories. Empirically, we observed that previous approaches failed to effectively address the problem of misclassification of videos from similar categories. Taking the action of the \textit{high jump} and \textit{long jump} as an instance, some methods (e.g., OTAM~\cite{cao2020few}) is easy to confuse the two classes by assigning close prediction scores due to their similarity in scenes and sub-actions, as shown in Fig.~\ref{fig:fig1}(a).
 We have analyzed the main reasons from three folds. (i) Class prototype construction: task-oriented class features can optimize videos' intra- and inter-class correlation. As shown in Fig.~\ref{fig:fig1}(b), most previous work has yet to use the whole task video features to extract relevant discriminative patterns. Although HyRSM~\cite{wang2022hybrid} manipulates interrelationship functions on different videos to get task-specific embeddings, it does not explicitly optimize intra- and inter-class correlations. (ii) Matching mechanisms: proper matching mechanisms need to be established to solve the confusion problem of similar videos. As shown in Fig.~\ref{fig:fig1}(c), current work almost all use a simple class prototype matching mechanism. Some methods use the frame-level matching mechanism~\cite{zhu2018compound, cao2020few, wang2022hybrid}, which is suitable for spatial-related datasets~\cite{
kuehne2011hmdb, soomro2012ucf101, carreira2017quo}, and the others use the tuple-level(multiple frames combined into a tuple) matching mechanism~\cite{thatipelli2022spatio, xing2023revisiting, perrett2021temporal} that is appropriate for temporal-related datasets~\cite{goyal2017something}. None of these previous methods can cope with video tasks of variable types well. (iii) Feature modeling: a powerful and highly discriminative feature is first needed to distinguish similar classes. Most previous works model the temporal feature through hand-designed temporal alignment algorithms ~\cite{zhu2018compound,cao2020few} or simple temporal attention operations~\cite{wang2022hybrid,xing2023revisiting}, leading to a simplistic exploration of the temporal relationship without dissecting it into more detailed patch and channel temporal relations to analyze.

Based on the above observations, we propose a novel method for few-shot action recognition, dubbed \textbf{GgHM}, a short for \textbf{G}raph-\textbf{g}uided \textbf{H}ybrid \textbf{M}atching. Specifically, we apply a graph neural network (GNN) for constructing task-oriented features, as shown in Fig.\ref{fig:fig1}(b). It could interactively transfer information between video features in a task to enhance the prior knowledge of the unknown video. We utilize the ground truth of the constructed graph edges to explicitly learn the correlation of these video features to supervise the similarity score learning between the query and support videos.  
Second, as shown in Fig.\ref{fig:fig1}(c), we propose a hybrid prototype matching strategy that combines frame-level and tuple-level matching based on the bidirectional Hausdorff Distance. Although the Hausdorff metric frame-level matching can alleviate the strictly ordered constraints of acquiring better query-support correspondences, it fails to capture temporal order. As a result, it can be confused for actions with similar action scenes strongly dependent on temporal order, e.g., \textit{putting something in the box} and \textit{taking something out of it}. However, the construction of tuples strictly follows a chronological order, which can compensate for the frame-level matching problem. Fig.\ref{fig:fig1}(a) visualizes the predicted similarities between query and support videos with different methods on the 5-way 1-shot task of UCF101~\cite{soomro2012ucf101}. Our method achieves more discriminative results for similar videos in each task compared to OTAM~\cite{cao2020few} and TRX~\cite{perrett2021temporal}.
Additionally, we design a learnable dense temporal modeling module to consolidate the representation foundation. It includes a temporal patch and temporal channel relation modeling block, and their combination allows for dense temporal modeling in both spatial and channel domains. Finally, extensive experiments on four widely-used datasets demonstrate the effectiveness of our method.

In summary, we make the following contributions:
\begin{itemize}
   \item We apply a graph neural network to guide the task-oriented features learning during the class prototype construction, explicitly optimizing the intra- and inter-class correlation within video features.
    \item We propose a hybrid class prototype matching strategy based on the frame- and tuple-level prototype matching, giving rise to effectively coping with video tasks of multivariate styles.
    \item We design a learnable dense temporal modeling module consisting of a temporal patch and temporal channel relation modeling block for dense temporal modeling in both spatial and channel domains.
\end{itemize}

\section{Related Works}
\subsection{Few-shot Image Classification}
	Few-shot image classification uses the episodic training paradigm, using a handful of labeled training samples from similar tasks to represent a large amount of labeled training samples. Recent years, research on few-shot image classification can be mainly classified into two categories: adaptation-based and metric-based methods. The adaption-based approaches aim to find a network initialization that can be fine-tuned for unknown tasks using a small amount of labeled data, called \textit{gradient by gradient}. The classical adaptation-based approaches are MAML~\cite{finn2017model}, Reptile~\cite{nichol2018reptile}, and related deeper researches include ~\cite{li2017meta, wang2020m}. The metric-based approaches aim to learn a feature space and compare task features through different matching strategies, called \textit{learning to compare}. The representative methods are Prototypical Networks~\cite{snell2017prototypical}, Matching Networks~\cite{vinyals2016matching}. And there are many methods~\cite{yoon2019tapnet, ye2020few, doersch2020crosstransformers, li2019finding} that aim to improve upon these approaches. Our method is inspired by them and belongs to the metric-based category.
 
    \subsection{Few-shot Video Action Recognition}
    The core idea of few-shot action recognition is similar to that of few-shot image classification, but the former task is more complex than the latter owning to an additional temporal dimension. Due to high computational resources and long experimental time, adaptation-based methods( MetaUVFS~\cite{patravali2021unsupervised}) have received little attention in few-shot action recognition. The existing research mainly applies metric-based learning, but with different focuses. Some methods focus on feature representation enhancement. For example, STRM~\cite{ thatipelli2022spatio} employs local and global enrichment modules for spatiotemporal modeling, HyRSM~\cite{wang2022hybrid} uses the hybrid relation modeling to learn task-specific embeddings, and SloshNet~\cite{xing2023revisiting} utilizes a feature fusion architecture search module to exploit the low-level spatial features and a long-term and short-term temporal modeling module to encode complementary global and local temporal representations. Other methods focus on class prototype matching strategies. For example, OTAM~\cite{cao2020few} proposes a temporal alignment module to calculate the distance value between the query video and the support set videos, TRX~\cite{perrett2021temporal} matches each query sub-sequence with all sub-sequences in the support set, HyRSM~\cite{ wang2022hybrid} designs a bidirectional Mean Hausdorff metric to more flexibly find the correspondences between different videos. Additionally, TRPN~\cite{wang2021semantic}, MORN~\cite{ni2022multimodal} focus on combining visual and semantic features, and AMeFu-Net~\cite{fu2020depth} centers on using depth information to assist learning. Unlike these previous methods, our method focuses on distinguishing videos from similar categories by optimizing intra- and inter-class class correlation within video features during the prototype construction and building a hybrid prototype matching strategy to effectively handle video tasks of multivariate styles.

\section{Method}
\subsection{Problem Formulation}
Few-shot learning is based on using a small number of labeled training samples from similar tasks as a proxy for many labeled training samples. For few-shot action recognition, it aims to classify an unlabeled query video into one of the $N$ action categories in the support set with limited $K$ samples per action class, which can be considered an $N$-way $K$-shot task. Like most previous studies, we adopt an episode training paradigm followed by~\cite{cao2020few, wang2022hybrid, huang2022compound,li2022hierarchical, xing2023revisiting}, where episodes are randomly selected from extensive data collection. In each episode, we suppose that the set $\mathcal{S}$ consists of $N\times~K$ samples from $N$ different action classes, and  $S_k^n=\left\{s_{k1}^n,s_{k2}^n,\cdots,s_{kT}^n\right\}$ represents the $k$-th video in class $n\ \in\left\{1,\cdots,N\right\}$ randomly sampled $T$ frames. The query video denotes $ Q=\left\{q_1,q_2,\cdots,q_T\right\}$ sampled $T$ frames.
\subsection{Architecture Overview}
\begin{figure*} [ht]
		\centering
		\includegraphics[width=\linewidth]{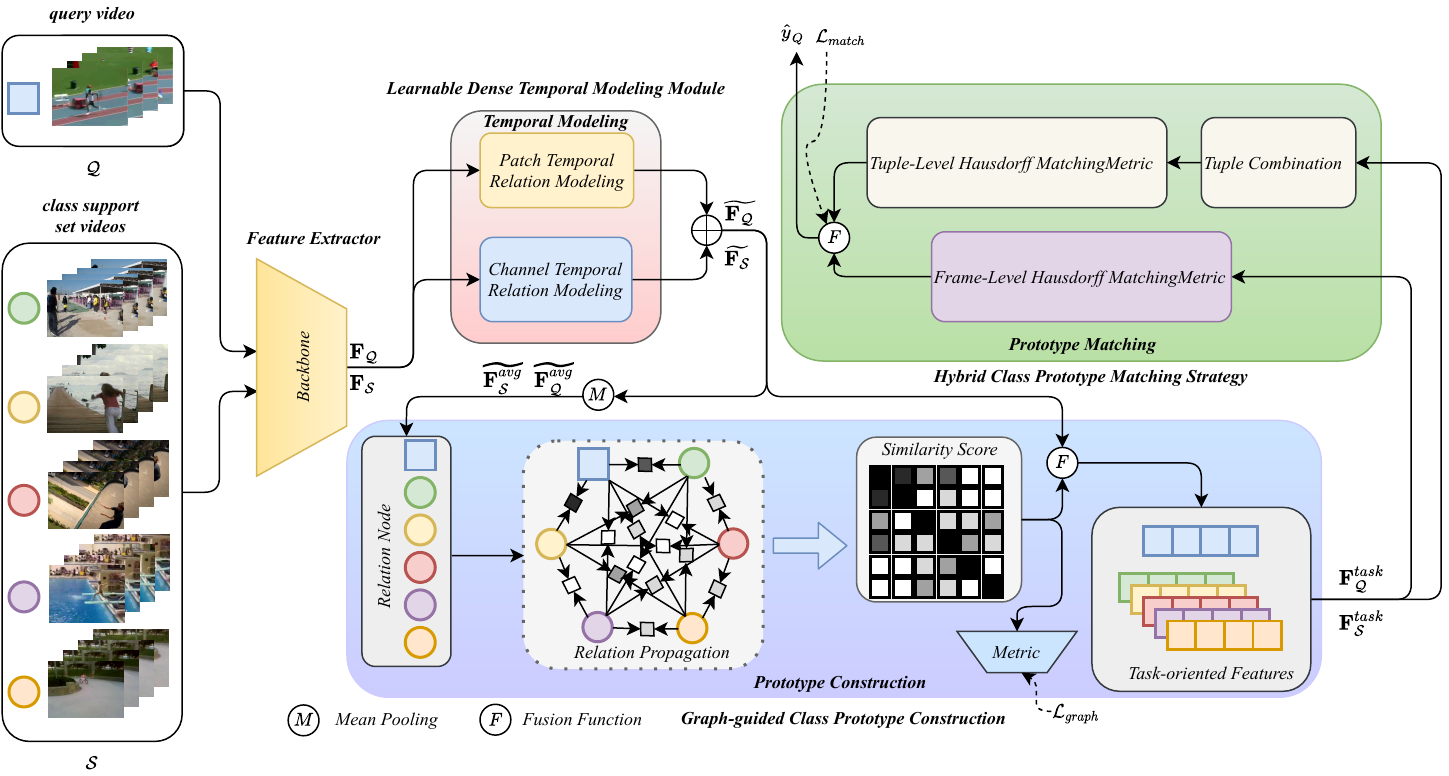}
	\caption{Overview of GgHM. For simplicity and convenience, we discuss the case: the 5-way 1-shot problem and the query set $\mathcal{Q}$ with a single video. The support set video features $\textbf{F}_\mathcal{S}$ and query video feature $\textbf{F}_\mathcal{Q}$ are obtained by the feature extractor. The enhanced temporal features $\widetilde{\textbf{F}_\mathcal{S}}$ and $\widetilde{\textbf{F}_\mathcal{Q}}$ are obtained by the learnable dense temporal modeling module. The  task-oriented features ${\textbf{F}_\mathcal{S}}^{task}$and ${\textbf{F}_\mathcal{Q}}^{task}$ are obtained by the graph-guided prototype construction module. The $\widehat{y}_Q$ is the class prediction of the query video, and the loss $\mathcal{L}_{match}$ and  $\mathcal{L}_{graph}$ are the standard cross-entropy loss. $\oplus$ indicates element-wise weighted summation.} 
		\label{fig:pipeline}
\vspace{-5pt}

\end{figure*}
Our overall architecture is illustrated in Fig.\ref{fig:pipeline}. For the frame-selecting strategy, we follow previous work TSN~\cite{wang2016temporal}, where the input video sequence is divided into $T$ segments, and snippets are extracted from each segment. For simplicity and convenience, we discuss the process of the 5-way 1-shot problem and consider that the query set $\mathcal{Q}$ contains a single video.
 In this way, the query video $Q=\left\{q_1,q_2,\cdots,q_T\right\}$ and the class support set videos $S^n=\left\{s_{1}^n,s_{2}^n,\cdots,s_{T}^n\right\}$ ($S^n\ \in \ \mathcal{S}=\left\{S^1, S^2,\cdots, S^5\right\}$) pass through the feature extractor to obtain the query feature $\textbf{F}_\mathcal{Q}$ and the support features $\textbf{F}_{S^n}$($\textbf{F}_{S^n}\in\textbf{F}_\mathcal{S}$) in each episode. 
Next, we input $\textbf{F}_\mathcal{S}$ and $\textbf{F}_\mathcal{Q}$ to the proposed learnable dense temporal modeling module to obtain enhanced temporal features $\widetilde{\textbf{F}_\mathcal{S}}$ and $\widetilde{\textbf{F}_\mathcal{Q}}$. We apply to mean pooling operation on $\widetilde{\textbf{F}_\mathcal{S}}$ and $\widetilde{\textbf{F}_\mathcal{Q}}$ in the temporal dimension to obtain the relation node features $\widetilde{\textbf{F}^{avg}_\mathcal{S}}$ and $\widetilde{\textbf{F}^{avg}_\mathcal{Q}}$ for the following graph network. Then, the relation node features are taken into the graph network with initial edge features for relation propagation. The updated edge features with enhanced temporal features generate task-oriented features ${\textbf{F}_\mathcal{S}^{task}}$ and ${\textbf{F}_\mathcal{Q}^{task}}$ and obtain the loss $\mathcal{L}_{graph}$ through a graph metric. Finally, the task-oriented features are fed into the hybrid class prototype matching metric to get the class prediction $\widehat{y}_Q$ and loss $\mathcal{L}_{match}$. 

For better clarity and consistency with the algorithm procedure, we will first introduce our learnable dense temporal modeling module, followed by the graph-guided prototype construction, and finally the hybrid prototype matching strategy.
Details are shown in the subsequent subsections.

\subsection{Learnable Dense Temporal Modeling Module (LDTM)}
\begin{figure} [ht]
		\centering
	\includegraphics[width=\linewidth,height=0.55\linewidth]{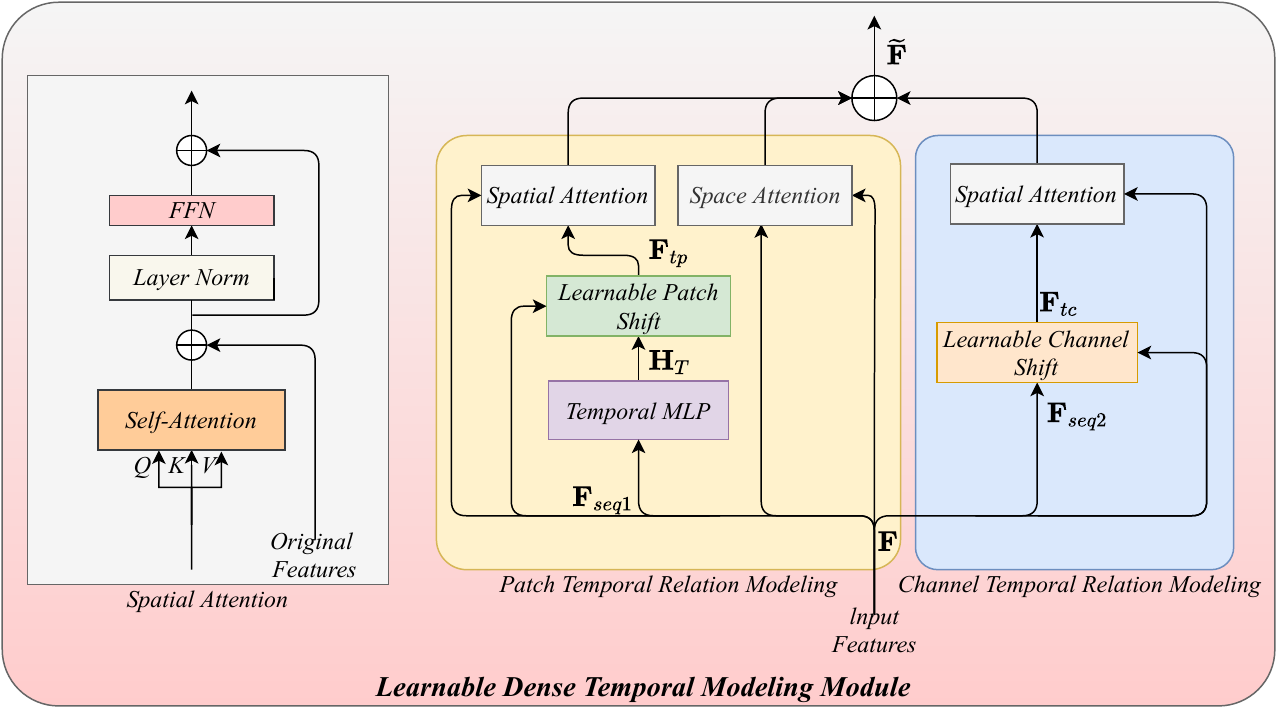}
		\caption{The architecture of learnable dense temporal modeling module. $\oplus$ denotes element-wise summation.}
		\label{fig:tp}
\vspace{-9pt}
	\end{figure}

The action classification process relies heavily on temporal context information. 
Inspired by some temporal modeling methods based on attention mechanism ~\cite{arnab2021vivit, zhang2021vidtr, fan2021multiscale, xiang2022spatiotemporal, bertasius2021space}, we design a learnable dense temporal modeling module, which consists of a temporal patch relation modeling block and a temporal channel relation modeling block, as shown in Fig.\ref{fig:tp}. The two blocks are complementary, and their combination allows for dense temporal modeling in both the spatial and channel domains. Compared to PST~\cite{xiang2022spatiotemporal}, which uses a fixed patch shift strategy and a channel shift strategy, our learnable patch and channel temporal relation modeling enables the extraction of richer features.
 
\textbf{Patch Temporal Relation Modeling (PTRM).}
 Given a video feature map output by the feature extractor $\textbf{F}{\in\mathbb{R}}^{N\times T\times C\times H\times W}$, we first reshape it to a sequence as $\textbf{F}_{seq1}{\in\mathbb{R}}^{N\times HW \times C \times T}$ and then fed it into the temporal MLP to get hidden temporal feature 
 $\textbf{H}_{T}$:
    \begin{equation}\label{}
    \textbf{H}_{T} = relu\left(\textbf{W}_{t1}\textbf{F}_{seq1}\right)\textbf{W}_{t2}+ \textbf{F}_{seq1}
    \end{equation}
where $\textbf{W}_{t1}$ and $\textbf{W}_{t2}{\in\mathbb{R}}^{T \times T}$ are the learnable weights for temporal information interaction of different video frames. Then, $\textbf{H}_{T}$  with rich video spatiotemporal information are inserted into the original features $\textbf{F}_{seq1}$, making the single-frame video feature contain semantic information for all video frames. The temporal patch relation modeling feature $\textbf{F}_{tp}$ is obtained by:
    \begin{equation}\label{}
   \textbf{F}_{tp}\left[\ :,\ n,\ :\ ,\ :\right] = \\
   \left\{ \begin{array}{c}
	\textbf{F}_{seq1}\left[\ :,\ n,\ :\ ,\ :\right] \,\, if\,\,n\%gap=0\\
	\textbf{H}_{T}\left[\ :,\ n,\ :\ ,\ :\right] \,\, if\,\,n\%gap \neq 0\\
    \end{array} \right. 
    \end{equation}
where $n$ is the patch index and $gap$ is a positive integer to control the frequency of the patch shift. After the learnable patch shift operation, the feature $\textbf{F}_{tp}$ is reshaped as $\textbf{F}_{tp}^{*}{\in\mathbb{R}}^{NT\times HW\times C}$ and do spatial self attention. This way collects the temporal information of the different video frames sparsely within the frame but sacrifices the original spatial information within every frame. To alleviate this problem, we do the weighted summation between spatial-only and spatiotemporal attention results, given by:
 \begin{equation}\label{}
\textbf{F}_{tp} = \gamma SA_{spa}\left(\textbf{F}_{tp}^{*}\right) + \left(\textbf1-\gamma\right)SA_{spa}\left(\textbf{F}^{*}\right)
\end{equation}
where $SA_{spa}$ stands for the spatial attention operation, $\textbf{F}^{*}{\in\mathbb{R}}^{NT\times HW\times C}$ is reshaped from $\textbf{F}$ and $\gamma\in\left[0,1\right]$ is a hyperparameter.

\textbf{Channel Temporal Relation Modeling (CTRM).}
We first reshape $\textbf{F}$ as $\textbf{F}_{seq2}{\in\mathbb{R}}^{NHW \times C \times T}$. Then it is fed it into a learnable channel shift operation to obtain the temporal channel relation modeling feature $\textbf{F}_{tc}$. Concretely, the learnable channel shift operation is a 1D channel-wise temporal convolution adopted to learn independent kernels for each channel. Formally, the learnable channel shift operation can be formulated as:
\begin{equation}\label{}
\textbf{F}_{tc}^{t,c} = \sum_{i}\textbf{K}_{c,i}\textbf{F}_{seq2}^{c,t+i}
\end{equation}
where $t$ and $c$ denote the temporal and channel dimensions of the feature map, respectively. $\textbf{K}_{c,i}$ indicates the temporal kernel weights of the $c$-th channel, $\textbf{F}_{seq2}^{c,t+i}\in\textbf{F}_{seq2}$ is the input $c$-th channel feature and  $\textbf{F}_{tc}^{t,c}\in\textbf{F}_{tc}$ is the output $c$-th channel feature.  After that, the final temporal channel relation modeling feature $\textbf{F}_{tc}$ is obtained through a spatial attention and we do the weight summation between $\textbf{F}_{tp}$ and  $\textbf{F}_{tc}$ to obtain the final enhanced temporal features $\widetilde{\textbf{F}}$ as follows:
 \begin{equation}\label{}
\widetilde{\textbf{F}} = \beta\textbf{F}_{tp} + \left(\textbf1-\beta\right)\textbf{F}_{tc} 
\end{equation}
where $\beta\in\left[0,1\right]$ is a hyperparameter.

In summary, PTRM aggregates temporal information for parts of patches while CTRM learns the temporal shift of channels. As a result, our LDTM could achieve sufficient temporal relation modeling in both the spatial and channel dimensions in a dense and learnable way.

\subsection{Graph-guided Prototype Construction(GgPC)}
We design a graph-guided prototype construction module to enhance the priori knowledge of the unknown video and explicitly optimize the intra- and inter-class correlation within video features. We draw inspiration from few-shot image classification methods based on graph neural networks~\cite{garcia2017few, kim2019edge, ma2020transductive, chen2021eckpn}, which utilize graph networks to optimize intra-cluster similarity and inter-cluster dissimilarity and transform the image classification problems into node or edge classification problems. Different from this, directly feeding the video features (usually after the temporal pooling operation) into the graph network can lead to unsatisfactory results due to the loss of temporal information. Therefore, we only use graph networks as guidance to optimize features' intra- and inter-class correlation.

\begin{figure} [t]
		\centering
	\includegraphics[width=\linewidth,height=0.65\linewidth]{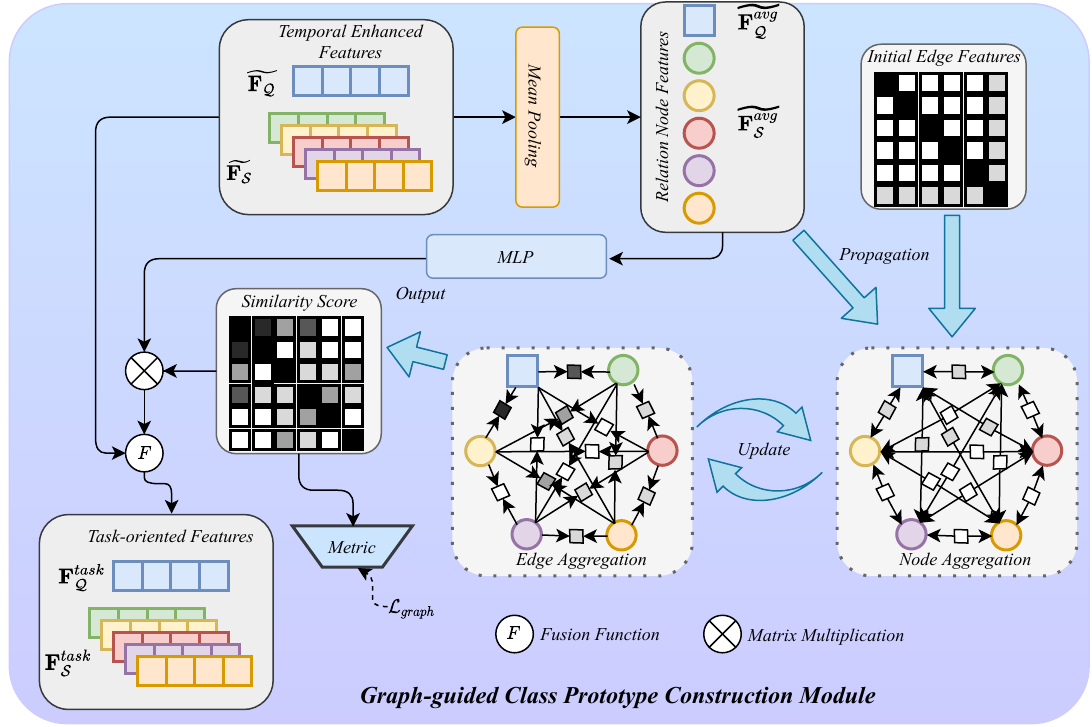}
		\caption{The overall framework of the proposed graph-guided prototype construction model. Consider that the query set $\mathcal{Q}$ contains one video for simplicity and convenience.}
		\label{fig:graph}
   \vspace{-9pt}

\end{figure}

The overall framework of the proposed graph-guided prototype construction module is shown in Fig.\ref{fig:graph}, and the overall algorithm is summarized in Algorithm.\ref{algorithm: algorithm1}. For simplicity and convenience, we discuss the process of the $N_S$-way 1-shot problem and consider that the query set $\mathcal{Q}$ contains $N_\mathcal{Q}$ videos. This process can be divided into two stages: Graph neural network (GNN) propagation and task-oriented features obtaining. For GNN propagation, the temporally enhanced features $\widetilde{\textbf{F}}$  after doing the Mean Pooling operation in the temporal dimension $\widetilde{\textbf{F}^{avg}}$ are used as node features $\textbf{V}$ for graph network initialization. Edge features $\textbf{A}$ represent the relationship between two nodes, i.e., the strength of intra- and inter-class relationships, and their initialization depends on the labels. The propagation includes the node aggregation and edge aggregation process. After completing the graph propagation, we use a $\tt Select$ operation to extract the similarity score from the updated edge features in the last layer. $\tt Select$ means that the edge features related to each query video feature are selected from the output entire edge features, and a total of $N_\mathcal{Q}$ new edge features are formed further.  For task-oriented features obtaining, the details are shown in Algorithm.\ref{algorithm: algorithm1} where $f_{FNN}$ is a feed-forward network, $f_{emb}$ and $f_{fuse}$ are MLPs, and $\otimes$ indicates the matrix multiplication. Meanwhile, the $\tt Select$ process is summarized in Algorithm.\ref{algorithm: algorithm2}. For $K$-shot ($K>1$) tasks, when constructing node features, we perform mean pooling on the features of support videos of the same category in the feature dimension, while keeping other aspects consistent with the 1-shot task.

To sum up, the task-oriented features ${\textbf{F}}^{task}$ are obtained by fusing enhanced temporal features $\widetilde{\textbf{F}}$ with features ${\textbf{F}}^{graph}$ guided by graph networks to preserve the temporality of features. Through the guidance of GNN, every query video feature has its special support features, and the class correlation within video features is optimized explicitly. 

\begin{algorithm}[t]
\label{algorithm: algorithm1}

\caption{The process of graph-guided prototype construction(GgPC)}
$\textbf{Us}$ indicates the $unsqueeze$ operation, $\textbf{R}$ indicates the $repeat$ operation.

{\bf Input:} $\widetilde{\textbf{F}_\mathcal{S}}{\in\mathbb{R}}^{N_\mathcal{S}\times T\times C}$, $\widetilde{\textbf{F}_\mathcal{Q}}{\in\mathbb{R}}^{N_\mathcal{Q}\times T\times C}$, $\widetilde{\textbf{F}}=\widetilde{\textbf{F}_\mathcal{S}}\bigcup\widetilde{\textbf{F}_\mathcal{Q}}=\widetilde{\textbf{F}}{\in\mathbb{R}}^{(N_\mathcal{S}+N_\mathcal{Q})\times T\times C}$

{\bf Output:} ${\textbf{F}}_\mathcal{Q}^{task}{\in\mathbb{R}}^{N_\mathcal{Q}\times T\times C}, {\textbf{F}}_\mathcal{S}^{task}{\in\mathbb{R}}^{N_\mathcal{Q}\times N_S\times T\times C}$

{\bf Initialize:}  $\widetilde{\textbf{F}^{avg}}={\tt Mean\_pool}(\widetilde{\textbf{F}},dim=1)$


\tcc{GNN Propagation}
{\bf Graph:} $\textbf{G}=\left(\textbf{V},\textbf{A};\mathcal{S} \bigcup \mathcal{Q}\right)$
, ${\bf v}_{i}^0 =\widetilde{\textbf{F}^{avg}_i}$, $\textbf{a}_{ij}^{0}$, $~\forall i,j\in{\mathcal{S}\bigcup \mathcal{Q}}$
    
    \For{$ l = 1,\cdots,L$}{
        \For{$i = 1,\cdots, |{\bf V}|$}{
    ${\bf v}_{i}^{l} = {\tt NodeAggregation} ({\bf v}_{j}^{l-1}, {\bf a}_{ij}^{l-1})$\
    }
        \For{$(i,j) = 1,\cdots, |\bf A|$}{
    ${\bf a}_{ij}^{l} = {\tt EdgeAggregation} ({\bf v}_{j}^{l}, {\bf a}_{ij}^{l-1})$\
    }
}
{\bf Similarity Score:} $\textbf{M}_{siam}={\tt Select}(\textbf{a}_{ij}^{L}[0]){\in\mathbb{R}}^{N_\mathcal{Q}\times (N_\mathcal{S}+1)\times (N_\mathcal{S}+1)}$ 

\tcc{Get Task-Oriented Features}

{\bf Optimized Features:} $\textbf{F}_\mathcal{S}^{node}= \widetilde{\textbf{F}^{avg}_\mathcal{S}}.\textbf{Us}(0).\textbf{R}(N_\mathcal{Q},1,1)$
$\textbf{F}^{node}={\tt Cat}([\textbf{F}_\mathcal{S}^{node},\widetilde{\textbf{F}^{avg}_\mathcal{Q}}.\textbf{Us}(1)], dim=1)$
${\textbf{F}}^{graph}= f_{FFN}(\textbf{M}_{siam}\otimes f_{emb}(\textbf{F}^{node})$)
${\textbf{F}_S}^{graph}= {\textbf{F}}^{graph}[:,:N_S,:].\textbf{Us}(1).\textbf{R}(1,T,1,1)$
${\textbf{F}_Q}^{graph}= {\textbf{F}}^{graph}[:,N_S:,:].\textbf{Us}(1).\textbf{R}(1,T,1)$

{\bf Task-oriented Features:} 
$\textbf{F}_S^{hid} = \widetilde{\textbf{F}_S}.\textbf{Us}(0).\textbf{R}(N_Q,1,1,1)$
${\textbf{F}_\mathcal{S}}^{task}= f_{fuse}({\tt Cat}([\textbf{F}_\mathcal{S}^{hid},{\textbf{F}_\mathcal{S}^{graph}}],dim=2))$
${\textbf{F}_\mathcal{Q}}^{task}= f_{fuse}({\tt Cat}([\widetilde{\textbf{F}_\mathcal{Q}},{\textbf{F}_\mathcal{Q}^{graph}},dim=2]))$
\end{algorithm}
\vspace{-5pt}

\begin{algorithm}[t]
\label{algorithm: algorithm2}

\caption{The process of  $\tt Select$ operation}
\label{Select}

{\bf Input:} $\textbf{a}_{ij}^{L}[0]{\in\mathbb{R}}^{(N_\mathcal{S}+N_\mathcal{Q})\times (N_\mathcal{S}+N_\mathcal{Q})}$

{\bf Output:} $\textbf{M}_{siam}{\in\mathbb{R}}^{N_\mathcal{Q}\times (N_\mathcal{S}+1)\times (N_\mathcal{S}+1)}$

{\bf Similarity Score:} 
$\textbf{M}_{siam} = {\tt List()}$\\
\For{$ n_\mathcal{Q} = 1,\cdots,N_\mathcal{Q}$}{
    $\textbf{m}_{siam} = {\tt Zeros}((N_\mathcal{S}+1)\times (N_\mathcal{S}+1))$\\
    $\textbf{m}_{siam}[:N_\mathcal{S},:N_\mathcal{S}] = \textbf{a}_{ij}^{L}[0][:N_\mathcal{S},:N_\mathcal{S}] $\\
    $\textbf{m}_{siam}[:N_\mathcal{S},-1] = \textbf{a}_{ij}^{L}[0][:N_\mathcal{S},N_\mathcal{S}+n_\mathcal{Q}] $\\
    $\textbf{m}_{siam}[-1,:N_\mathcal{S}] = \textbf{a}_{ij}^{L}[0][N_\mathcal{S}+n_\mathcal{Q},:N_\mathcal{S}] $\\
    $\textbf{m}_{siam}[-1,-1] = \textbf{a}_{ij}^{L}[0][N_\mathcal{S}+n_\mathcal{Q},N_\mathcal{S}+n_\mathcal{Q}] $\\
    $\textbf{M}_{siam}.{\tt Append}(\textbf{m}_{siam})$
    }
$\textbf{M}_{siam} = {\tt Stack}(\textbf{M}_{siam})$
\end{algorithm}

\subsection{Hybrid Prototype Matching Strategy (HPM)}
Frame-level matching uses single individual frames, while tuple-level matching combines several frames into a tuple as the matching unit. HyRSM~\cite{wang2022hybrid} applies the Hausdorff Distance metric as the prototype matching method, which can alleviate the strictly ordered constraints of acquiring better query-support correspondences, but it fails to capture temporal order. This matching metric is easily confused for actions with similar action scenes but strongly depends on temporal order,e.g., $pick\ up\ a\ glass\ of\ water$ and $put\ down\ a\ glass\ of\ water$. To solve this problem, we design a hybrid prototype matching strategy that combines frame-level and tuple-level matching based on the bidirectional Hausdorff Distance. This approach effectively copes with video tasks of diverse styles. Given the task-oriented features $\textbf{F}_\mathcal{S}^{task}$, $\textbf{F}_\mathcal{Q}^{task}$, the $m$-th support video feature in the $k$ class and the $p$-th query video feature indicates $\textbf{s}^k_m{\in\mathbb{R}}^{T \times C}$, $\textbf{q}_p{\in\mathbb{R}}^{T \times C}$, respectively. For single-frame matching, we apply a bidirectional Mean Hausdorff metric as follow:
 \begin{equation}\label{}
 \begin{split}
   \mathcal{D}_{frame} = \frac{1}{T} \Bigg[\sum_{\textbf{s}^k_{m,i}\in\textbf{s}^k_m}{\left( \min_{\textbf{q}_{p,j}\in\textbf{q}_p} \!\:\left\| \textbf{s}^k_{m,i}- \textbf{q}_{p,j}\right\| \right)} \\+\sum_{\textbf{q}_{p,j}\in\textbf{q}_p}{\left( \min_{\textbf{s}^k_{m,i}\in\textbf{s}^k_m} \!\:\left\| \textbf{q}_{p,j} -\textbf{s}^k_{m,i} \right\| \right)}\Bigg]
\end{split}
\end{equation}
where $\textbf{s}^k_{m,i}$ represents the $i$-th frame feature of $\textbf{s}^k_m$, $\textbf{q}_{p,j}$ indicates the $j$-th frame feature of $\textbf{q}_p$, and they have a total $T$ frames. For tuple-level prototype matching, we combine two frames into one tuple and iterate through all combinations to get $L=\frac{1}{2}\left( T-1 \right)T$ tuples for $T$ frames, given by:
 \begin{equation}\label{}
 \begin{split}
 \textbf{ts}^k_{m,i} = \left[\textbf{s}^k_{m,i_1}+\textbf{PE}(i_1),\textbf{s}^k_{m,i_2}+\textbf{PE}(i_2)\right] \ 1\leqslant i_1\leqslant i_2\leqslant T \\
\textbf{tq}_{p,j} = \left[\textbf{q}_{p,j_1}+\textbf{PE}(j_1),\textbf{q}_{p,j_2}+\textbf{PE}(j_2)\right] \ 1\leqslant j_1\leqslant j_2\leqslant T
\end{split}
\end{equation}
where $\textbf{ts}^k_{m,i},\textbf{tq}_{p,j}{\in\mathbb{R}}^{2C}$, and each tuple follows the temporal information of the original frame. To this end, the Mean Hausdorff metric based on tuples can be formulated as:
 \begin{equation}\label{}
 \begin{split}
   \mathcal{D}_{tuple} = \frac{1}{L} \Bigg[\sum_{\textbf{ts}^k_{m,i}\in\textbf{ts}^k_m}{\left( \min_{\textbf{tq}_{p,j}\in\textbf{tq}_p} \!\:\left\| \textbf{ts}^k_{m,i}- \textbf{tq}_{p,j}\right\| \right)} \\+\sum_{\textbf{tq}_{p,j}\in\textbf{tq}_p}{\left( \min_{\textbf{ts}^k_{m,i}\in\textbf{ts}^k_m} \!\:\left\| \textbf{tq}_{p,j} -\textbf{ts}^k_{m,i} \right\| \right)}\Bigg]
\end{split}
\end{equation}
Finally, the hybrid matching metric can be formulated as:
\begin{equation}\label{}
\mathcal{D}_{hybrid} = \alpha\mathcal{D}_{tuple} + \left(\textbf1-\alpha\right)\mathcal{D}_{frame}
\end{equation}
where $\alpha\in\left[0,1\right]$ is a hyperparameter.

In a word, our proposed hybrid prototype matching strategy combines the advantages of both frame- and tuple-level matching to cope with video tasks of multivariate styles well.

\section{Experiments}
\subsection{Experimental Setup} 
\textbf{Datasets.} We evaluate the performance of our method on four few-shot datasets, including Kinetics~\cite{carreira2017quo}, HMDB51~\cite{kuehne2011hmdb}, UCF101~\cite{soomro2012ucf101}, and SSv2~\cite{goyal2017something}. For Kinetics and SSv2, we use the splits provided by \cite{cao2020few} and \cite{zhu2020label}, where 100 classes were selected and divided into 64/12/24 action classes as the meta-training/meta-validation/meta-testing set. Additionally, for UCF101 and HMDB51, we evaluate our method on the splits provided by ~\cite{zhang2020few}.
\begin{table*}[t]
\centering
\begin{tabular}{c|c|c|cc|cc|cc|cc}
\hline
\rowcolor[HTML]{ECF4FF} & & &
\multicolumn{2}{c|}{HMDB51} & \multicolumn{2}{c|}{UCF101} & \multicolumn{2}{c|}{SSv2} & \multicolumn{2}{c}{Kinetics} \\
\rowcolor[HTML]{ECF4FF}\multirow{-2}{*}{Methods} & \multirow{-2}{*}{Reference}&  \multirow{-2}{*}{Backbone}&1-shot & 5-shot & 1-shot & 5-shot & 1-shot & 5-shot & 1-shot & 5-shot \\ \hline
MatchingNet~\cite{vinyals2016matching} &NeurIPS(16)& ResNet-50 &- & - & 31.3 & 45.5 & - & - & 53.3 & 74.6 \\
MAML~\cite{finn2017model} &ICML(17)& ResNet-50&- & - & 30.9 & 41.9 & - & - & 54.2 & 75.3 \\ 
ProtoNet~\cite{snell2017prototypical}& NeurIPS(17)&C3D& 54.2 & 68.4 & 74.0 & 89.6 & 33.6 & 43.0 & 64.5 & 77.9 \\
TRN++~\cite{zhou2018temporal} & ECCV(18)& ResNet-50 & - & - & - & - & 38.6 & 48.9 & 68.4 & 82.0 \\
CMN++~\cite{zhu2018compound} & ECCV(18)& ResNet-50 &- & - & - & 34.4 & 43.8 & - & 57.3 & 76.0 \\
TARN~\cite{bishay2019tarn} & BMVC(19)&  C3D& - & - & - & - & - & - & 64.8 & 78.5 \\
ARN ~\cite{zhang2020few} & ECCV(20)&  C3D&45.5 & 60.6 & 66.3 & 83.1 & - & - & 63.7 & 82.4 \\
OTAM~\cite{cao2020few} & CVPR(20)& ResNet-50 & 54.5 & 68.0 & 79.9 & 88.9 & 42.8 & 52.3 & 73.0 & 85.8 \\
TTAN~\cite{li2021ttan}& ArXiv(21)& ResNet-50 & 57.1 & 74.0 & 80.9 & 93.2 & 46.3 & 60.4 & - & - \\
ITANet~\cite{zhang2021learning} & IJCAI(21)& ResNet-50 & - & - & - & - & 49.2 & 62.3 & 73.6 & 84.3 \\
TRX~\cite{perrett2021temporal} & CVPR(21)& ResNet-50 & 54.9* & 75.6 & 81.0* & 96.1 & 42.0 & 64.6 & 65.1* & 85.9 \\
TA2N~\cite{li2022ta2n} & AAAI(22)& ResNet-50 &59.7 &  73.9 & 81.9 & 95.1 & 47.6 & 61.0 & 72.8 & 85.8\\
STRM~\cite{thatipelli2022spatio}  & CVPR(22)& ResNet-50 & 57.6* & \underline{77.3} & 82.7* & 96.9 & 43.5* & 66.0* & 65.1* & 86.7 \\
MTFAN~\cite{wu2022motion} & CVPR(22)& ResNet-50 & 59.0 & 74.6 & 84.8 & 95.1 & 45.7 & 60.4 & \underline{74.6} & \textbf{87.4} \\
HyRSM~\cite{wang2022hybrid} & CVPR(22)& ResNet-50 & \underline{60.3} & 76.0 & 83.9 & 94.7 & \underline{51.5*} & 67.5* & 73.7 & 86.1 \\
HCL~\cite{zheng2022few} & ECCV(22)& ResNet-50 & 59.1 & 76.3 & 82.5 & 93.9 & 47.3 & 64.9 & 73.7 & 85.8 \\
Huang $et al.$~\cite{huang2022compound}& ECCV(22)& ResNet-50 & 60.1 & 77.0 & 71.4 & 91.0 & 49.3 & 66.7 & 73.3 & 86.4 \\
Nguyen $et al.$~\cite{nguyen2022inductive}& ECCV(22)& ResNet-50 & 59.6 & 76.9 & 84.9 & 95.9 & 43.8 & 61.7 & 74.3 & \textbf{87.4} \\
SloshNet~\cite{xing2023revisiting} & AAAI(23)& ResNet-50 & 59.4 & \textbf{77.5} & \textbf{86.0} & \textbf{97.1} & 46.5 & \underline{68.3} & 70.4 & \underline{87.0} \\
\hline
 \textbf{GgHM} & - & ResNet-50& \textbf{61.2} & 76.9 & \underline{85.2}& \underline{96.3} & \textbf{54.5} & \textbf{69.2} & \textbf{74.9} & \textbf{87.4} \\ \hline
\end{tabular}
\caption{State-of-the-art comparison on the 5-way k-shot benchmarks of HMDB51, UCF101, SSv2, Kinetics. The \textbf{boldfacen} and \underline{underline font} indicate the highest and the second highest results.  Note: * means our implementation.}
\label{tab:compare}
\vspace{-9pt}
\end{table*}
\textbf{Network Architectures.} We utilize the ResNet-50 as the feature extractor with ImageNet pre-trained weights~\cite{deng2009imagenet}. For LDTM, $\textbf{W}_{t1}, \textbf{W}_{t2}$ are two one-layer MLPs, and $gap$ is set to 2. For GgPC, we apply one-layer GNN to obtain task-oriented features. More implementation details can be found in the appendix.

     \textbf{Training and Inference.} Followed by TSN~\cite{wang2016temporal}, we uniformly sample 8 frames ($T$=8) of a video as the input augmented with some basic methods, e.g. random horizontal flipping, cropping, and color jit in training, while multi-crops and multi-views in inference. For training, SSv2 were randomly sampled 100,000 training episodes, and the other datasets were randomly sampled 10,000 training episodes. Moreover, we used the Adam optimizer with the multi-step scheduler for our framework. For inference, we reported the average results over 10,000 tasks randomly selected from the test sets in all datasets. 
\subsection{Results}

 As shown in Tab.\ref{tab:compare}, our method \textbf{GgHM} achieves impressive results against the state-of-the-art methods in all datasets and few-shot settings. Our method especially achieves new state-of-the-art performance on Kinetics and SSv2 in all few-shot settings and HMDB in the 5-way 1-shot task, respectively. In other tasks, our method either achieves the second-highest result or achieves results that are very close to the SOTA. Our method performs impressively without any preference for datasets or the few-shot settings.
In contrast, some methods perform unsatisfactorily in the 1-shot task (e.g., TRX~\cite{perrett2021temporal}, STRM~\cite{thatipelli2022spatio}, SloshNet~\cite{xing2023revisiting}) or particular datasets (e.g., Nguyen $et al.$~\cite{nguyen2022inductive} on SSv2, MTFAN~\cite{wu2022motion} on SSv2, Huang $et al.$~\cite{huang2022compound} on UCF101). In addition, compared to our baseline HyRSM~\cite{wang2022hybrid}, which also utilizes the Hausdorff Distance metric as the class prototype matching strategy and focuses on building the task-oriented feature, the effect of our method is significantly improved. Specifically, compared to HyRSM, our method brings 0.9$\%$, 1.3$\%$, 3.0$\%$, and 0.3$\%$  performance improvements in the 1-shot task of HMDB51, UCF101, SSv2, and Kinetics, respectively. In the 5-shot task, our method outperforms HyRSM significantly, bringing 0.3$\%$, 1.6$\%$, 2.7$\%$, and 0.7$\%$ gain on HMDB51, UCF101, SSv2, and Kinetics, respectively.

\subsection{Ablation Study}

\textbf{Impact of the proposed components.} To validate the contributions of each module (i.e. LDTM, GgPC, HPM) in our method, we experiment under 5-way 1-shot and 5-way 5-shot settings on the SSv2 dataset. Our baseline method only utilizes the frame-level bidirectional Mean Hausdorff metric as the  prototype matching strategy without any extra modules. As shown in Tab.~\ref{proposed modules}, we observe that each component is effective. Specifically, compared to the baseline, the HPM module can bring 0.6$\%$ and 0.7$\%$ accuracy improvement on 1-shot and 5-shot tasks, the GgPC module can bring 4.4$\%$ and 5.5$\%$ performance improvement on two tasks, and the LDTM module can bring 7.2$\%$ and 8.9$\%$ performance gain on two tasks. Additionally, stacking modules can enhance performance, indicating the complementarity between components. Combining all modules can get the best results, bringing 9.9$\%$ and 13.2$\%$ performance improvement on 1-shot and 5-shot tasks over the baseline.
\begin{table}[t!]
\centering
\scalebox{0.95}{\begin{tabular}{c|c|c|c|c}
\hline
\rowcolor[HTML]{FFCCC9}
    LDTM & GgPC  & HPM & 1-shot & 5-shot \\ \hline
    $\usym{2717}$ 	 &  $\usym{2717}$     & $\usym{2717}$   & 44.6 & 56.0 \\ \hline  
    $\usym{2717}$ 	 &  $\usym{2717}$     & $\usym{2713}$ & 45.2& 56.7 \\ \hline
    $\usym{2717}$ 	 &  $\usym{2713}$ & $\usym{2717}$   & 49.0 & 61.5 \\ \hline  
    $\usym{2713}$ &  $\usym{2717}$     & $\usym{2717}$   & 51.8 & 64.9\\ \hline
    $\usym{2717}$     &  $\usym{2713}$ &$\usym{2713}$   & 50.1 & 63.4\\ \hline
     $\usym{2713}$ 	 &  $\usym{2717}$     & $\usym{2713}$   & 52.2 & 65.8 \\ \hline
    $\usym{2713}$     &  $\usym{2713}$ &$\usym{2717}$   & 53.9 & 68.7\\ \hline
    $\usym{2713}$    &  $\usym{2713}$    &   $\usym{2713}$& \textbf{54.5} & \textbf{69.2} \\ \hline
\end{tabular}}
\caption{The impact of proposed modules on SSv2 in the 5-way 1-shot and 5-way 5-shot settings.} 
\label{proposed modules}
\vspace{-9pt}
\end{table}

\textbf{Impact of temporal modeling integration.}  To explore the impact of each temporal modeling module in LDTM and demonstrate their effectiveness, we experiment on the 5-way 1-shot and 5-way 5-shot tasks of SSV2 to ablate our proposed temporal relation modeling blocks. The PTRM block includes spatial attention, which indicates doing Self-Attention only on the spatial dimension.  As shown in Tab.\ref{Temporal}, the CTRM block brings about a 1.0$\%$ and 1.9$\%$ accuracy improvement on the 1-shot and 5-shot tasks over the baseline. Moreover, the PTRM block obtains 1.5$\%$ and 2.3$\%$ gain on the 1-shot and 5-shot tasks over the baseline. The integration of these two blocks results in 2.9$\%$ and 3.7$\%$ gain on two tasks, respectively. 

\begin{table}[b!]
\centering
\scalebox{0.95}{\begin{tabular}{c|c|c|c|c}
\hline
\rowcolor[HTML]{FFCCC9}
    Spatial Attention  & PTRM  & CTRM & 1-shot & 5-shot \\ \hline
   $\usym{2713}$ 	 &  $\usym{2717}$     & $\usym{2717}$   & 51.6 &65.5 \\ \hline  
    $\usym{2713}$	  &  $\usym{2717}$     & $\usym{2713}$  & 52.6 & 67.4\\ \hline
    $\usym{2717}$  &  $\usym{2713}$       &$\usym{2717}$   & 53.1 & 67.8\\ \hline
$\usym{2717}$  &  $\usym{2713}$    &   $\usym{2713}$& \textbf{54.5} & \textbf{69.2} \\ \hline
\end{tabular}}
\caption{The impact of temporal modeling blocks integration on SSv2 in the 5-way 1-shot and 5-way 5-shot settings.} 
\label{Temporal}

\end{table}
 
\textbf{Analysis of building the task-oriented features.} To demonstrate the necessity of constructing task-specific features and compare the efficacy of various methods for constructing them, we conduct experiments on the 5-way 1-shot task of Kinetics and SSv2. Building task-oriented features can be divided into two categories: unsupervised and supervised. The critical difference between them is whether label information is used directly to constrain the construction of features. The Self-Attention method(HyRSM~\cite{wang2022hybrid}) means that the task features (the set of support and query video features) do self-attention without using the label information to supervise. In contrast, our GNN method directly applies label information to do supervision, which can explicitly optimize the video features' intra-
and inter-class correlation. As shown in Tab.\ref{Graph}, the Self-Attention method can bring 1.2$\%$ and 1.5$\%$ gain on Kinetics and SSv2 over the baseline each, which can demonstrate the necessity of building task-oriented features. Moreover, our GNN method(each query feature owns a graph) can bring 1.7$\%$ and 1.8$\%$ gain over the baseline on two datasets, respectively, showing the advantage of the supervised method. Moreover, our GNN method with transduction(all query features in the same graph) brings a 2.0$\%$ and 2.3$\%$ accuracy improvement on two datasets. 
\begin{table}[t!]
\centering
\scalebox{0.95}{\begin{tabular}{c|c|c|c}
\hline
\rowcolor[HTML]{FFCCC9}
      Method& Type  & Kinetics  & SSv2   \\ \hline
        None & -   & 72.9       & 52.2  \\ \hline 
  Self-Attention& unsupervised& 74.1      & 53.7 \\ \hline  
    GNN&	supervised  &  74.6    & 54.0\\ \hline
    GNN(Transduction)& supervised & \textbf{74.9} & \textbf{54.5} \\ \hline
\end{tabular}}
\caption{Analysis of building the task-oriented features on Kinetics and SSv2 in the 5-way 1-shot setting.} 
\label{Graph}
\end{table}

\begin{table}[t!]
\centering
\scalebox{0.95}{\begin{tabular}{c|c|c}
\hline
\rowcolor[HTML]{FFCCC9}
      Metric        & Kinetics  & SSv2    \\ \hline
        Frame-level matching     & 74.3       & 53.9  \\ \hline 
  Tuple-level matching & 74.1      & 54.2 \\ \hline  
    Hybrid matching	  &  \textbf{74.9}    & \textbf{54.5}\\ \hline
\end{tabular}}
\caption{Comparisons of different prototype matching strategies on Kinetics and SSv2 in the 5-way 1-shot setting.} 
\label{Matchings}

\end{table}

\begin{table}[t!]
\centering
\scalebox{0.95}{\begin{tabular}{c|c|c|c|c|c|c}
\hline
\rowcolor[HTML]{FFCCC9}
      Param $\alpha$        & 0  & 0.2 &0.4& 0.6& 0.8 & 1.0    \\ \hline
     Kinetics     & 74.3& 74.6 & \textbf{74.9} & 74.5 & 74.3 &74.1  \\ \hline 
        SSv2     & 53.9 & 54.1 & 54.2 & \textbf{54.5} & 54.3 &54.2 \\ \hline  
\end{tabular}}
\caption{The impact of the varying fusion parameter $\alpha$ of hybrid prototype matching on Kinetics and SSv2 in the 5-way 1-shot setting.  }
\label{alpha}
\vspace{-9pt}

\end{table}

 \begin{figure} [b!]
    \centering
		\includegraphics[width=\linewidth,height=0.56\linewidth]{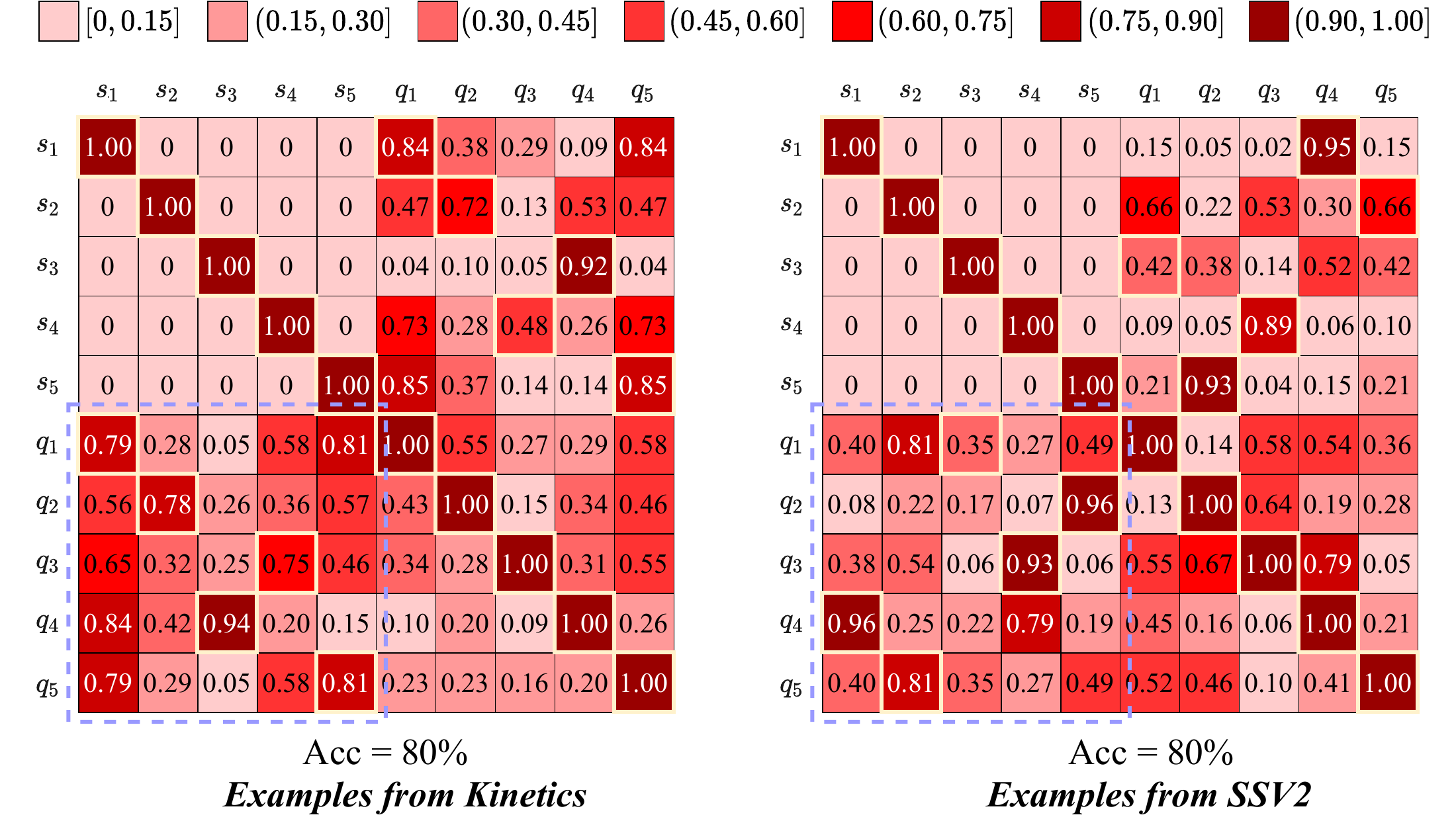}
		\caption{Visualization of the updated edge features output by the GNN. $GT$ stands for the ground truth and $ACA$ represents the accuracy calculation area. A higher score indicates a greater degree of similarity. We can use the features in the accuracy calculation area directly to obtain task recognition results.}
		\label{fig:siam_score}

	\end{figure}

 \begin{figure*} [th!]
		\centering
		\includegraphics[width=\linewidth,height=0.46\linewidth ]{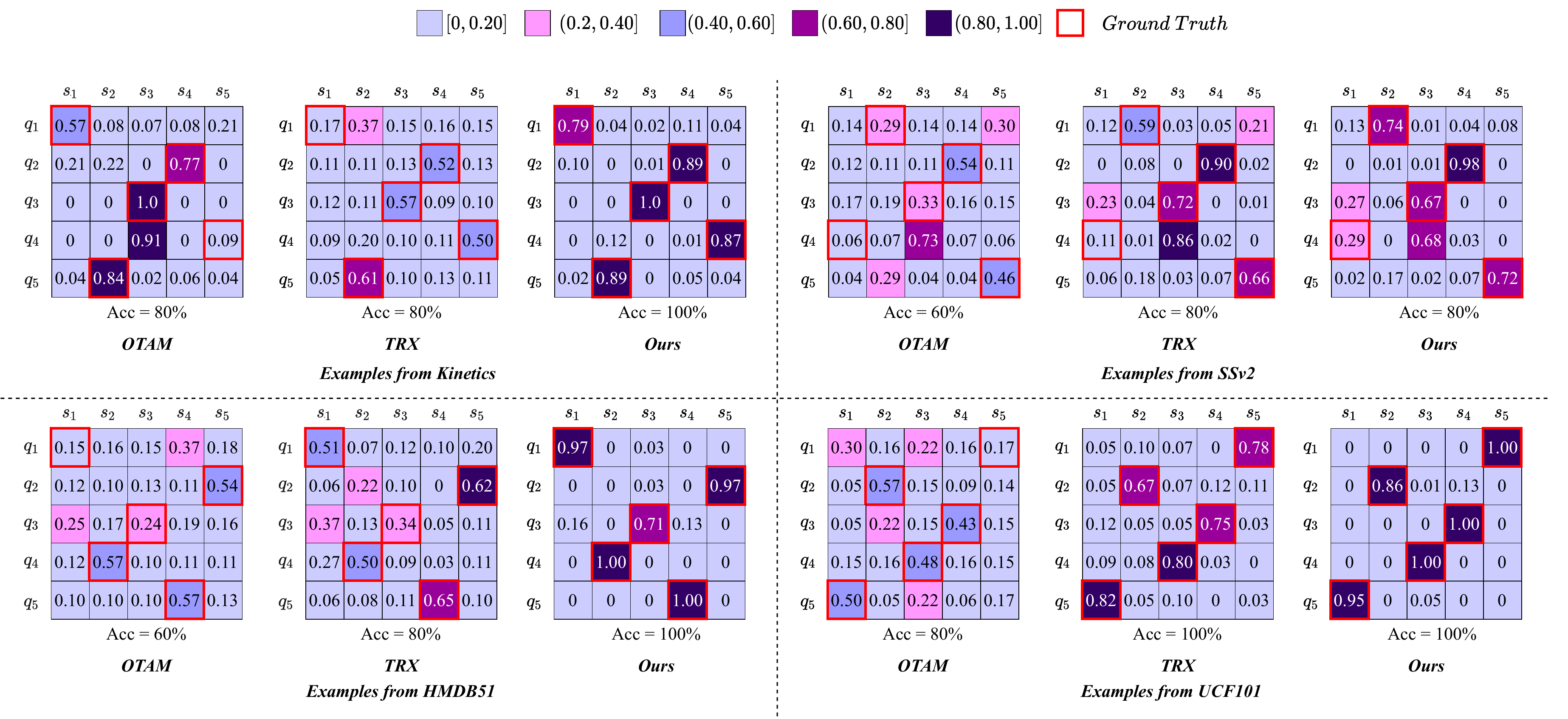}
	\caption{ Similarity visualization between query and support videos with different methods on the 5-way 1-shot task of Kinetics, SSv2, HMDB51, and UCF101. A higher score indicates a greater degree of similarity.} 
		\label{fig:similarity}
		\vspace{-9pt}
\end{figure*}
\textbf{Comparisons of different prototype matching strategies.} To analyze different prototype matching strategies, we experiment on the 5-way 1-shot task of Kinetics and SSv2 with different  prototype matching methods to evaluate the effectiveness of our hybrid matching strategy. All the methods are based on the bidirectional Mean Hausdorff metric and the experiment results are shown in Tab.\ref{Matchings}. Our hybrid matching strategy brings a  0.6$\%$ and   0.6$\%$ accuracy improvement on two datasets over the frame-level matching strategy. Meanwhile, it obtains 0.8$\%$ and 0.3$\%$ gain on two datasets over the tuple-level matching strategy, respectively.

\textbf{Impact of the varying fusion parameter of hybrid prototype matching.} Tab.\ref{alpha} shows the impact of the varying fusion parameter $\alpha$ in hybrid prototype matching. As part of our experiments, we perform the 5-way 1-shot task on Kinetics and SSV2. The parameter $\alpha$ denotes the weight assigned to the frame- and tuple-level matching in the final fusion. From the results, the optimal values of parameter $\alpha$ are 0.4 for Kinetics and 0.6 for SSv2.

\textbf{Visualization of the update edge features output by GNN.} As shown in Fig.\ref{fig:siam_score}, we visualize two examples of the updated edge features output by the GNN and the ground truth on Kinetics and SSv2 in the 5-way 1-shot setting. The edge features' value can be seen as the similarity score between two video features. From the visualization, GNN as guidance can well optimize video features' inter- and intra-class correlation, in which updated edge features are very close to the similarity matrix corresponding to the ground truth. Meanwhile, the intermediate output recognition results of GNN obtained by the edge features in the accuracy calculation area can also achieve high accuracy.

\textbf{Similarity visualization.}
Fig.\ref{fig:similarity} visualizes the predicted similarities between query and support videos with different methods on the 5-way 1-shot task of Kinetics, SSv2, HMDB51, and UCF101. Our method achieves more discriminative results for similar videos in each task compared to OTAM~\cite{cao2020few} and TRX~\cite{perrett2021temporal}. The results presented here demonstrate the effectiveness of our method in distinguishing videos from similar categories, as it has significantly improved both the prediction accuracy and intra-/inter-class correlation within video features.

\section{Conclusion}
In this work, we have presented a novel few-shot action recognition framework, GgHM, leading to impressive performance in recognizing similar categories in every task without any datasets or task preference. Concretely, we learn task-oriented features by the guidance of a graph neural network during class prototype construction, optimizing the intra- and inter-class feature correlation explicitly. Next, we propose a hybrid class prototype matching strategy that leverages both frame- and tuple-level prototype matching to effectively handle video tasks with diverse styles. Besides, we propose a dense temporal modeling module consisting of a temporal patch and temporal channel relation modeling block to enhance the video feature temporal representation, which helps to build a more solid foundation for the matching process. GgHM shows consistent improvements over other challenging baselines on several few-shot datasets, demonstrating the effectiveness of our method.

\section*{Acknowledgement}
This work is partly supported by the following grant: Key
R\&D Program of Zhejiang (No.2022C03126).

\section*{Supplementary Materials}
\subsection*{Details on GNN Propagation in GgPC}

 Graph neural networks(GNN) are well established for the application~\cite{garcia2017few, kim2019edge, ma2020transductive, chen2021eckpn} of few-shot image classification. In our method, we followed  EGNN~\cite{kim2019edge} to utilize GNN as guidance to optimize the intra- and inter-class correlation within features. For simplicity and convenience, we discuss the process of the $N_\mathcal{S}$-way 1-shot problem and consider that the query set $\mathcal{Q}$ contains $N_\mathcal{Q}$ videos.
We let  $\textbf{G}=\left(\textbf{V},\textbf{A};\mathcal{S} \bigcup \mathcal{Q}\right)$ be the graph to construct the relationship between support set videos $\mathcal{S}$ and query videos $\mathcal{Q}$. We use the video features as node features $\textbf{V}=\left\{\textbf{v}_i\right\}_{i=1,\cdots,\left|\mathcal{S} \bigcup \mathcal{Q}\right|}$ and the relationship between the node features as edge features $\textbf{A}=\left\{\textbf{a}_{ij}\right\}_{i,j=1,\cdots,\left|\mathcal{S} \bigcup \mathcal{Q}\right|}$, where $\left|\mathcal{S} \bigcup \mathcal{Q}\right|= N_\mathcal{S}+N_\mathcal{Q}$. 

Node features are initialized by the enhanced temporal features after the mean pooling operation on the temporal dimension, i.e., ${\bf v}_{i}^0 =\widetilde{\textbf{F}^{avg}_i}(\forall i\in{\mathcal{S}\bigcup \mathcal{Q}})$. Edge features $\textbf{a}_{ij}{\in\mathbb{R}}^{2}(\forall i,j\in{\mathcal{S}\bigcup \mathcal{Q}})$ are 2D vectors representing the intra- and inter-class relations of the two  connected nodes and are initialized with ground-truth $y$, as follows:
 \begin{equation}\label{}
 \begin{split}
 \textbf{a}_{ij}^{0}=\left\{\begin{array}{cc} {[1||0]},& \mbox{if} ~ y_{i} = y_{j} ~~ \mbox{and}~~ i, j \le N_{\mathcal{S}}, \\ {[0||1]},&  \mbox{if} ~ y_{i}\ne y_{j} ~~ \mbox{and}~~ i, j \le N_{\mathcal{S}}, \\ {[0.5||0.5]},& \mbox{otherwise}, 
\end{array} \right. 
\end{split}
\end{equation}
The $\textbf{G}$ consists of $L$ layers, and its propagation includes node features and edge features updating. Given $\textbf{v}_i^{l-1}{\in\mathbb{R}}^{C}$ and $\textbf{a}_i^{l-1}{\in\mathbb{R}}^{2}$ from the layer $l-1$, node features' updating is a weighted aggregation process of other nodes through the  layers' edge features, as follows:
\begin{equation}
{\bf v}_{i}^{l} = f_{node}^{l}\big({\tt Cat}\big([\sum_{j} {\tilde a}_{ij1}^{l-1}{\bf v}_j^{l-1},\sum_{j} {\tilde a}_{ij2}^{l-1}{\bf v}_j^{l-1}], dim=0\big)\big)
\label{eq: node update}
\end{equation}
where $f_{node}^{l}$ is a MLP to transform feature and ${\tilde a}_{ijb}^{l-1} = \frac{a_{ijb}^{l-1}}{\sum_{h}a_{ihb}^{l-1}}$ on $b\in\{1,2\}$. After the update of node features, the edge feature is updated through the (dis)similarities between two connected features, and the sum of all edge features' values is kept constant, given by:
\begin{equation}
{\bar a_{ijb}^{l}}=\left\{ \begin{array}{c}
	\frac{f_{edge}^{l}\left( \left| \textbf{v}_i^l-\textbf{v}_j^l \right| \right) a_{ijb}^{l-1}}{\sum_h{f\left( \left| \textbf{v}_i^l-\textbf{v}_h^l \right| \right)  a_{ihb}^{l-1}}}\sum_ha_{ihb}^{l-1}, \,\,\,\     if\,\,b=0\\
 \\
	\frac{
 \left(1-f_{edge}^{l}\big( \left| \textbf{v}_i^l-\textbf{v}_j^l \right| \right)\big) a_{ijb}^{l-1}}{\sum_h{\big(1-{f_{edge}^{l}\left( \left| \textbf{v}_i^l-\textbf{v}_h^l \right| \right) \big)}  a_{ihb}^{l-1}}}\sum_ha_{ihb}^{l-1},\,\,\,\   if\,\,b=1\\
\end{array} \right. 
\label{eq: edge update}
\end{equation}
\begin{equation}
{\bf a}_{ij}^{l} = {\bar {\bf a}}_{ij}^{\l} / \|{\bar {\bf a}}_{ij}^{l}\|_1
\end{equation}
where $f_{edge}^{l}$ is a function to calculate the similarities between two connected nodes.  Here we set $f_{edge}^{l}$ to a  four-layer convolution block, where each layer comprises a 1$\times$1 convolutional layer, batch normalization, and LeakyReLU activation function. 

\begin{figure*} [t]
		\centering
		\includegraphics[width=\linewidth]{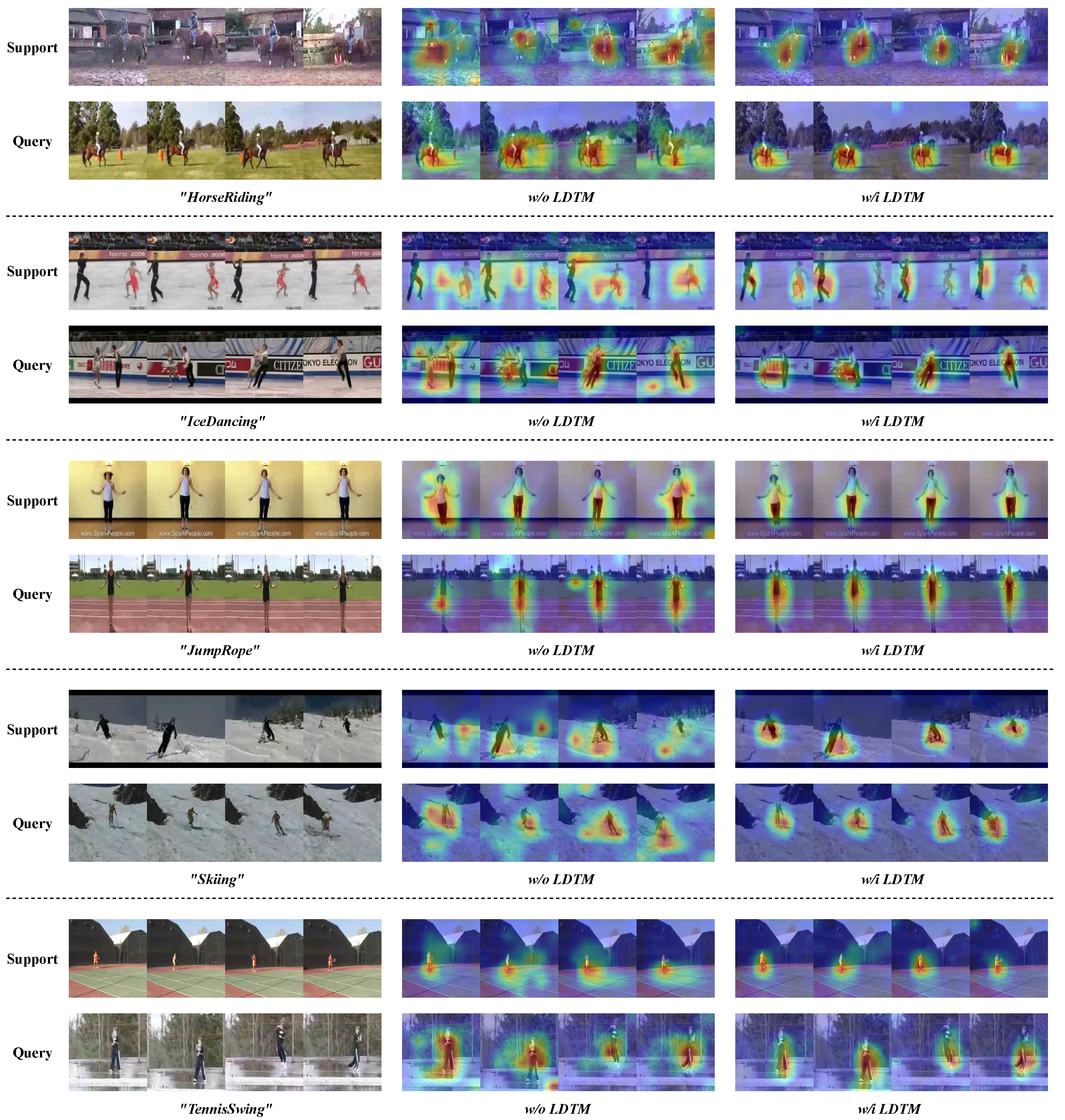}
	\caption{ Attention visualization of our GgHM on UCF101 in the 5-way 1-shot setting. Corresponding to the original RGB images (left), the attention maps without LDTM modules (middle) are compared to the attention maps with our LDTM modules (right).} 
		\label{fig:attn_ucf}
  \vspace{-10pt}
\end{figure*}
\subsection*{Implementation Details of Experimental Setup}
\subsubsection*{Network Architectures}
The kernel size for the 1D channel-wise temporal convolution in CTRM is set to 3. The settings of hyperparameters in each dataset are shown in Tab.\ref{hyperparameters}.
\begin{table}[th]
\centering
\scalebox{0.95}{\begin{tabular}{c|c|c|c|c}
\hline
              & Kinetics  & SSv2 & UCF101 & HMDB51  \\ \hline
 $\gamma$     & 0.1      & 0.5    &0.1     &0.1   \\ \hline 
 $\beta$      & 0.9      & 0.5    &0.9     &0.9 \\ \hline  
 $\alpha$	 & 0.4      &0.6     &0.5     &0.5   \\ \hline
\end{tabular}}
\caption{The settings of hyperparameters in each dataset.} 
\label{hyperparameters}
  \vspace{-3pt}

\end{table}

\subsubsection*{Training and Inference}
In HPM, when $T$ is set to 8, $L$ is calculated as 32. The total number of training steps is set to 10. Tab.\ref{Training} presents the learning rate and other settings for various datasets. In this table, $lr$ refers to the learning rate, $st\_iter$ indicates the number of iterations per step, $steps$ represents the number of steps to change the learning rate when using the multi-step scheduler, and $LRS$ denotes the multiplication factor for updating the learning rate at each changing step.

\begin{table}[th]
\centering
\scalebox{0.90}{\begin{tabular}{c|c|c|c|c}
\hline
              &$lr$  &  $st\_iter$ & $steps$ &  $LRS$  \\ \hline
Kinetics      & 2.2e-5      &  1000     &[0,6,9]    &[1,0.5,0.1]   \\ \hline 
 SSv2 &  1e-4      & 7500  &[0,6,8,9]     &[1,0.5,0.1,0.01]   \\ \hline 
HMDB51    &  1e-4      & 1000    & [0,2,3,5]  &[1,0.5,0.1,0.01] \\ \hline  
UCF101	 & 5e-05     &1500     &[0,2,3,5]    &[1,0.5,0.1,0.01]   \\ \hline

\end{tabular}}
\caption{The settings of hyperparameters in each dataset.} 
\label{Training}
\vspace{-3pt}

\end{table}


\subsection*{Attention Visualization of our GgHM}
Fig.\ref{fig:attn_ucf} shows the attention visualization of our GgHM on UCF101 in the 5-way 1-shot setting. Compared to the original RGB images on the left, the attention maps without LDTM modules (in the middle) are contrasted against the attention maps with our LDTM modules (on the right). Attention maps generated without the LDTM module contain numerous irrelevant or distracting focus areas. For example,  the frames in ``$HorseRiding$" show attention to the background and extraneous objects, diverting focus from the action. In contrast, attention maps generated using the LDTM module strongly correlate with the subject acting. Specifically, the frames in ``$Skiing$" focus on the skier, and the frames in ``$TennisSwing$" focus on the tennis player. These observations provide empirical evidence of the effectiveness of our LDTM module in enhancing spatiotemporal representation.
{\small
\bibliographystyle{ieee_fullname}
\bibliography{egbib}
}

\end{document}